\documentclass{clv2}
\usepackage{times}
\usepackage{url}
\usepackage{latexsym}
\usepackage{graphicx}
\usepackage{epsfig}
\usepackage{amsmath}
\usepackage{amssymb,CJK}
\usepackage{subfigure}
\newcommand{\argmax}{\operatornamewithlimits{argmax}}

\issue{1}{1}{2012}

\runningtitle{Probabilistic Models for High-Order Projective Dependency Parsing}

\runningauthor{Ma and Zhao}

\begin{document}
\title{Probabilistic Models for High-Order Projective Dependency Parsing}

\author{Xuezhe Ma\thanks{$^{1}$Center for Brain-Like Computing and Machine Intelligence
Department of Computer Science and Engineering. $^{2}$MOE-Microsoft Key Laboratory for Intelligent Computing and Intelligent Systems.
$\textrm{} \qquad \textrm{} \qquad \textrm{} \qquad \textrm{} \qquad \textrm{} \qquad$
Shanghai Jiao Tong University, 800 Dong Chuan Rd., Shanghai 200240, China.
$\textrm{} \qquad \textrm{} \qquad \textrm{} \qquad \textrm{} \qquad \textrm{} \qquad \textrm{} \qquad$
E-mail: xuezhe.ma@gmail.com, zhaohai@cs.sjtu.edu.cn}}
\affil{Shanghai Jiao Tong University}

\author{}
\affil{}

\author{Hai Zhao}
\affil{Shanghai Jiao Tong University}


\maketitle

\begin{abstract}
This paper presents generalized probabilistic models for high-order projective dependency parsing
and an algorithmic framework for learning these statistical models involving dependency trees.
Partition functions and marginals for high-order
dependency trees can be computed efficiently, by adapting our algorithms which extend the inside-outside
algorithm to higher-order cases. To show the effectiveness of our algorithms, we perform
experiments on three languages---English, Chinese and Czech, using maximum conditional likelihood estimation for model
training and L-BFGS for parameter estimation. Our methods achieve competitive performance for English,
and outperform all previously reported dependency parsers for Chinese and Czech.
\end{abstract}

\section{Introduction}
Dependency parsing is an approach to syntactic analysis inspired by dependency grammar.
In recent years, several domains of Natural Language Processing have benefited from dependency representations,
such as synonym generation~\cite{Shin:2002}, relation extraction~\cite{nguyen-moschitti-riccardi:2009:EMNLP} and
machine translation~\cite{katzbrown-EtAl:2011:EMNLP,xie-mi-liu:2011:EMNLP}. A primary reason for using dependency
structures instead of more informative constituent structures is that they are usually easier to be understood and
is more amenable to annotators who have good knowledge of the target domain but lack of deep linguistic
knowledge~\cite{Yamada:2003} while still containing much useful information needed in application.

Dependency structure represents a parsing tree as a directed graph with different
labels on each edge, and some methods based on graph models have been applied to it and achieved
high performance. Based on the report of the CoNLL-X shared task on dependency parsing~\cite{Buchholz:2006,Nivre:2007b},
there are currently two dominant approaches for data-driven dependency parsing: local-and-greedy
transition-based algorithms~\cite{Yamada:2003,Nivre:2004,Attardi:conll2006,McDonaldNivre:2007},
and globally optimized graph-based algorithms~\cite{eisn:1996,McDonald:2005,McDonald:2005b,McDonald:EACL06,cars:2007,Koo:2010},
and graph-based parsing models have achieved state-of-the-art accuracy for a wide range of languages.

There have been several existing graph-based dependency parsers, most of which employed online learning
algorithms such as the averaged structured perceptron~(AP)~\cite{FreundSchapire:1999,Collins:2002} or Margin
Infused Relaxed Algorithm~(MIRA)~\cite{Cram1:2003,Cram2:2003,McDonald:2006} for learning parameters.
However, One shortcoming of these parsers is that learning parameters of these models
usually takes a long time~(several hours for an iteration). The primary reason is
that the training step cannot be performed in parallel, since for online learning algorithms,
the updating for a new training instance depends on parameters updated with the previous instance.

Paskin~\shortcite{Paskin:2001} proposed a variant of the inside-outside algorithm~\cite{Baker:1979}, which were
applied to the grammatical bigram model~\cite{eisn:1996}. Using this algorithm, the grammatical bigram model can
be learning by off-line learning algorithms. However, the grammatical bigram model is based on a strong independence
assumption that all the dependency edges of a tree are independent of one another. This assumption restricts the model
to first-order factorization~(single edge), losing much of the contextual information in dependency tree.
Chen~et.al \shortcite{Chen:2010} illustrated that a wide range of decision history can lead to significant
improvements in accuracy for graph-based dependency parsing models. Meanwhile, several previous
works~\cite{cars:2007,Koo:2010} have shown that grandchild interactions provide important information for
dependency parsing. Therefore, relaxing the independence assumption for higher-order parts to capture much richer
contextual information within the dependency tree is a reasonable improvement of the bigram model.

In this paper, we present a generalized probabilistic model that can be applied to any types of factored models for
projective dependency parsing, and an algorithmic framework for learning these statistical models.
We use the grammatical bigram model as the backbone, but relax the independence assumption and
extend the inside-outside algorithms to efficiently compute the partition functions and
marginals (see Section~\ref{subsec:plb}) for three higher-order models.
Using the proposed framework, parallel computation technique can be employed,
significantly reducing the time taken to train the parsing models. To achieve empirical evaluations of our parsers,
these algorithms are implemented and evaluated on three treebanks---Penn WSJ Treebank~\cite{Marcus:1993} for English,
Penn Chinese Treebank~\cite{Xue:2005} for Chinese and Prague Dependency Treebank~\cite{Haj:1998,Haj:2001}
for Czech, and we expect to achieve an improvement in parsing performance. We also give an error analysis on structural
properties for the parsers trained by our framework and those trained by online learning algorithms.
A free distribution of our implementation has been put on the Internet.\footnote{\url{http://sourceforge.net/projects/maxparser/}}.

The remainder of this paper is structured as follows: Section~\ref{sec:dp} describes the probabilistic models and the
algorithm framework for training the models. Related work is presented in Section~\ref{sec:rw}. Section~\ref{sec:al}
presents the algorithms of different parsing models for computing partition functions and marginals.
The details of experiments are reported in Section~\ref{sec:pe}, and conclusions are in Section~\ref{sec:con}.

\begin{figure}[t]
\epsfig{figure=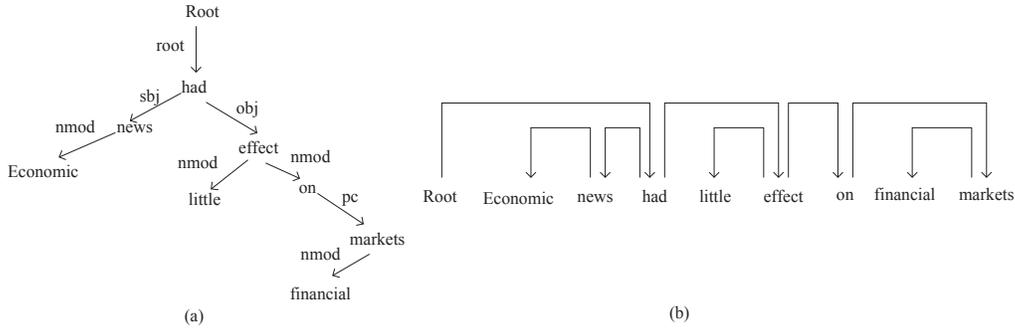,width=\textwidth}
\caption{An example dependency tree.}
\label{fig:dp-tree}
\end{figure}

\section{Dependency Parsing}
\label{sec:dp}
\subsection{Background of Dependency Parsing}
Dependency trees represent syntactic relationships through labeled directed edges of words and their syntactic
modifiers. For example, Figure~\ref{fig:dp-tree} shows a dependency tree for the sentence,
\emph{Economic news had little effect on financial markets}, with the sentence's root-symbol as its root.

By considering the item of crossing dependencies, dependency trees fall into two categories---projective and
non-projective dependency trees. An equivalent and more convenient formulation of the projectivity constrain is
that if a dependency tree can be written with all words in a predefined linear order and all edges drawn on the
plane without crossing edges~(see Figure~\ref{fig:dp-tree}(b)). The example in Figure~\ref{fig:dp-tree} belongs to the
class of projective dependency trees where crossing dependencies are not allowed.

Dependency trees are often typed with labels for each edge to represent additional syntactic information~(see
Figure~\ref{fig:dp-tree}(a)), such as $\mathrm{sbj}$ and $\mathrm{obj}$ for verb-subject and verb-object
head-modifier interactions, respectively. Sometimes, however, the dependency labels are omitted. Dependency trees are
defined as \emph{labeled} or \emph{unlabeled} according to whether the dependency labels are included or dropped.
In the remainder of this paper, we will focus on unlabeled dependency parsing for both theoretical and practical reasons.
From theoretical respect, unlabeled parsers are easier to describe and understand, and algorithms for unlabeled parsing can
usually be extended easily to the labeled case. From practical respect, algorithms of labeled parsing generally have higher
computational complexity than them of unlabeled version, and are more difficult to implement and verify.
Finally, the dependency labels can be accurately tagged by a \emph{two-stage labeling} method~\cite{McDonald:2006},
utilizing the unlabeled output parse.

\subsection{Probabilistic Model}
\label{subsec:pm}
The symbols we used in this paper are denoted in what follows, $\boldsymbol{x}$ represents a generic input sentence, and $\boldsymbol{y}$ represents a generic
dependency tree. $\mathrm{T}(\boldsymbol{x})$ is used to denote the set of possible dependency trees for sentence $\boldsymbol{x}$. The probabilistic model for dependency parsing defines a family of conditional probability $\textrm{Pr}(\boldsymbol{y}|\boldsymbol{x})$ over all $\boldsymbol{y}$ given sentence $\boldsymbol{x}$,
with a log-linear form:
\begin{displaymath}
\textrm{Pr}(\boldsymbol{y}|\boldsymbol{x})=\frac{1}{Z(\boldsymbol{x})}\exp
\bigg\{\sum_{j}\lambda_{j}F_{j}(\boldsymbol{y},\boldsymbol{x})\bigg\},
\end{displaymath}
where $F_{j}$ are feature functions, $\mathbf{\lambda} = (\lambda_{1}, \lambda_{2}, \ldots)$ are parameters of the model,
and $Z(\boldsymbol{x})$ is a normalization factor, which is commonly referred to as the \emph{partition function}:
\begin{displaymath}
Z(\boldsymbol{x})=\sum_{\boldsymbol{y} \in \mathrm{T}(\boldsymbol{x})}\exp \bigg\{\sum_{j}\lambda_{j}F_{j}(\boldsymbol{y},\boldsymbol{x})\bigg\}.
\end{displaymath}

\subsection{Maximum Likelihood Parameter Inference}
Maximum conditional likelihood estimation is used for model training (like a CRF). For a set of training
data $\{(\boldsymbol{x}_{k},\boldsymbol{y}_{k})\}$,
the logarithm of the likelihood, knows as the log-likelihood, is given by:
{\setlength\arraycolsep{2pt}
\begin{eqnarray}
L(\mathbf{\lambda}) & = & \log \prod\limits_{k} \textrm{Pr}(\boldsymbol{y}_{k}|\boldsymbol{x}_{k}) \nonumber \\
 & = & \sum\limits_{k}\log \textrm{Pr}(\boldsymbol{y}_{k}|\boldsymbol{x}_{k}) \nonumber \\
 & = & \sum\limits_{k}\bigg[ \sum\limits_{j}\lambda_{j}F_{j}(\boldsymbol{y}_{k},\boldsymbol{x}_{k}) - \log Z(\boldsymbol{x}_{k})\bigg]. \nonumber
\end{eqnarray}
}

Maximum likelihood training chooses parameters such that the log-likelihood $L(\mathbf{\lambda})$ is maximized.
This optimization problem is typically solved using quasi-Newton numerical methods such as \mbox{L-BFGS \cite{Nash:1991}},
which requires the gradient of the objective function:
{\setlength\arraycolsep{2pt}
\begin{eqnarray}\label{equ:gradient}
\frac{\partial L(\mathbf{\lambda})}{\partial \lambda_{j}} & = & \sum\limits_{k}\frac{\partial \log \textrm{Pr}(\boldsymbol{y}_{k}|\boldsymbol{x}_{k})}{\partial \lambda_{j}} \nonumber \\
 & = & \sum\limits_{k}\bigg[F_{j}(\boldsymbol{y}_{k},\boldsymbol{x}_{k}) - \frac{\partial \log
 z(\boldsymbol{x}_{k})}{\partial \lambda_{j}}\bigg] \\
 & = & \sum\limits_{k}\bigg[F_{j}(\boldsymbol{y}_{k},\boldsymbol{x}_{k}) -
\sum\limits_{y\in \mathrm{T}(x_{k})} \textrm{Pr}(\boldsymbol{y}|\boldsymbol{x}_{k})
F_{j}(\boldsymbol{y},\boldsymbol{x}_{k})\bigg]. \nonumber
\end{eqnarray}
}
The computation of $Z(\boldsymbol{x})$ and the second item in summation of
Equation (\ref{equ:gradient}) are the difficult parts in model training. In the following, we will
show how these can be computed efficiently using the proposed algorithms.

\subsection{Problems of Training and Decoding}
\label{subsec:plb}
In order to train and decode dependency parsers, we have to solve three inference problems which
are central to the algorithms proposed in this paper.

The first problem is the decoding problem of finding the best parse for a sentence when all
the parameters of the probabilistic model have been given. According to decision theory, a reasonable
solution for classification is the \emph{Bayes classifier} which classify to the most probable class,
using the conditional distribution. Dependency parsing could be regarded as a classification problem, so
decoding a dependency parser is equivalent to finding the dependency tree $\boldsymbol{y}^{*}$ which has
the maximum conditional probability:
{\setlength\arraycolsep{2pt}
\begin{eqnarray}\label{equ:decode}
\boldsymbol{y}^{*} & = & \argmax_{y\in \mathrm{T}(x)}\textrm{Pr}(\boldsymbol{y}|\boldsymbol{x}) \nonumber \\
 & = & \argmax_{y\in \mathrm{T}(x)}\log \textrm{Pr}(\boldsymbol{y}|\boldsymbol{x}) \nonumber \\
 & = & \argmax_{y\in \mathrm{T}(x)} \bigg\{ \sum\limits_{j}\lambda_{j}F_{j}(\boldsymbol{y},\boldsymbol{x}) \bigg\}.
\end{eqnarray}
}
The second and third problems are the computation of the partition function $Z(\boldsymbol{x})$ and the gradient of the log-likelihood (see Equation~(\ref{equ:gradient})).

From the definition above, we can see that all three problems require an exhaustive
search over $\mathrm{T}(\boldsymbol{x})$ to accomplish a maximization or summation. It is obvious
that the cardinality of $\mathrm{T}(\boldsymbol{x})$ grows exponentially with the length of $\boldsymbol{x}$,
thus it is impractical to perform the search directly. A common strategy is to \emph{factor}
dependency trees into sets of small \emph{parts} that have limited interactions:
\begin{equation}\label{equ:fts}
F_{j}(\boldsymbol{y},\boldsymbol{x})=\sum_{p\in y}f_{j}(p, \boldsymbol{x}).
\end{equation}
That is, dependency tree $y$ is treated as a set of parts $p$ and each feature function
$F_{j}(\boldsymbol{y},\boldsymbol{x})$ is equal to the sum of all the features $f_{j}(p, \boldsymbol{x})$.

We denote the \emph{weight} of each part $p$ as follows:
\begin{displaymath}
w(p,\boldsymbol{x})=\exp \bigg\{ \sum_{j}\lambda_{j}f_{j}(p,\boldsymbol{x}) \bigg\}.
\end{displaymath}
Based on Equation (\ref{equ:fts}) and the definition of weight for each part, conditional probability $\textrm{Pr}(\boldsymbol{y}|\boldsymbol{x})$ has the the following form:
\begin{eqnarray}
\textrm{Pr}(\boldsymbol{y}|\boldsymbol{x}) & = & \frac{1}{Z(\boldsymbol{x})}\exp
\bigg\{\sum\limits_{j}\lambda_{j}\sum\limits_{p\in y}f_{j}(p,\boldsymbol{x})\bigg\} \nonumber \\
 & = & \frac{1}{Z(\boldsymbol{x})}\exp
 \bigg\{\sum\limits_{p\in y}\sum\limits_{j}\lambda_{j}f_{j}(p,\boldsymbol{x})\bigg\} \nonumber \\
 & = & \frac{1}{Z(\boldsymbol{x})}\prod\limits_{p\in y} w(p,\boldsymbol{x}) \nonumber
\end{eqnarray}
Furthermore, Equation~(\ref{equ:decode}) can be rewritten as:
\begin{displaymath}
\boldsymbol{y}^{*}=\argmax_{y\in \mathrm{T}(x)}\sum_{p\in y}\log w(p,\boldsymbol{x}),
\end{displaymath}
and the partition function $Z(\boldsymbol{x})$ and the second item in the summation of Equation~(\ref{equ:gradient}) are
\begin{displaymath}
Z(\boldsymbol{x}) = \sum\limits_{y\in \mathrm{T}(x)}\bigg[\prod\limits_{p\in y} w(p,\boldsymbol{x})\bigg],
\end{displaymath}
and
{\setlength\arraycolsep{2pt}
\begin{eqnarray}
& & \sum\limits_{y\in \mathrm{T}(x_{k})} \textrm{Pr}(\boldsymbol{y}|\boldsymbol{x}_{k})
F_{j}(\boldsymbol{y},\boldsymbol{x}_{k}) \nonumber \\
 & = & \sum\limits_{y\in \mathrm{T}(x_{k})}\sum\limits_{p\in y}\textrm{Pr}(\boldsymbol{y}|\boldsymbol{x}_{k})
f_{j}(p,\boldsymbol{x}_{k}) \nonumber \\
 & = & \sum\limits_{p\in \mathrm{P}(x_k)}\sum\limits_{y\in \mathrm{T}(p, x_k)}
f_{j}(p,\boldsymbol{x}_{k})\textrm{Pr}(\boldsymbol{y}|\boldsymbol{x}_{k}) \nonumber \\
 & = & \sum\limits_{p\in \mathrm{P}(x_k)}f_{j}(p,\boldsymbol{x}_{k})\sum\limits_{y\in \mathrm{T}(p, x_k)}
\textrm{Pr}(\boldsymbol{y}|\boldsymbol{x}_{k}), \nonumber
\end{eqnarray}
}
where $\mathrm{T}(p,\boldsymbol{x})=\{\boldsymbol{y}\in \mathrm{T}(\boldsymbol{x})|p \in \boldsymbol{y}\}$
and $\mathrm{P}(x)$ is the set of all possible part $p$ for sentence $\boldsymbol{x}$.
Note that the remaining problem for the computation of the gradient in Equation~(\ref{equ:gradient})
is to compute the \emph{marginal probability} $m(p)$ for each part $p$:
\begin{displaymath}
m(p)=\sum\limits_{y\in \mathrm{T}(p,x)}\textrm{Pr}(\boldsymbol{y}|\boldsymbol{x}).
\end{displaymath}
Then the three inference problems are as follows:
\begin{itemize}
\item[] Problem 1: \textbf{Decoding}
\begin{displaymath}
\boldsymbol{y}^{*}=\argmax_{y\in \mathrm{T}(x)}\sum_{p\in y}\log w(p,\boldsymbol{x}).
\end{displaymath}
\item[] Problem 2: \textbf{Computing the Partition Function}
\begin{displaymath}
Z(\boldsymbol{x})=\sum\limits_{y\in \mathrm{T}(x)}\bigg[\prod\limits_{p\in y} w(p,\boldsymbol{x})\bigg].
\end{displaymath}
\item[] Problem 3: \textbf{Computing the Marginals}
\begin{displaymath}
m(p)=\sum\limits_{y\in \mathrm{T}(p,x)}\textrm{Pr}(\boldsymbol{y}|\boldsymbol{x}) \textrm{, for all }p.
\end{displaymath}
\end{itemize}

\subsection{Discussion}
It should be noted that for the parsers trained by online learning algorithms such as AP or MIRA, only the algorithm
for solving the decoding problem is required. However, for the motivation of training parsers using off-line parameter
estimation methods such as maximum likelihood described above, we have to carefully design algorithms for the inference
problem 2 and 3.

The proposed probabilistic model is capable of generalization to any types of parts $p$,
and can be learned by using the framework which solves the three inference problems. For different types of
factored models, the algorithms to solve the three inference problems are different.
Following Koo and Collins~\shortcite{Koo:2010}, the order of a part is defined as the number of dependencies
it contains, and the order of a factorization or parsing algorithm is the maximum of the order of the parts it uses.
In this paper, we focus on three factorizations: sibling and grandchild, two
different second-order parts, and grand-sibling, a third-order part:
\begin{figure}[h]
\centering
\epsfig{figure=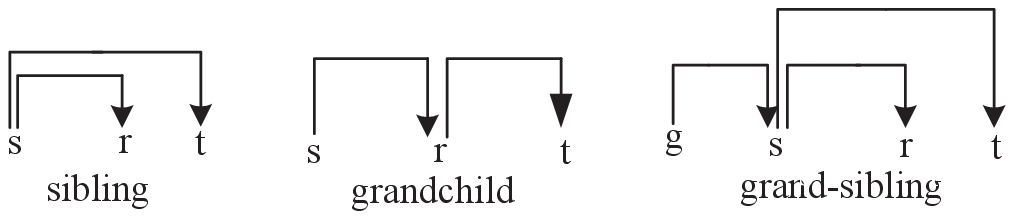,width=0.7\textwidth}
\end{figure}

In this paper, we consider only \emph{projective} trees, where crossing
dependencies are not allowed, excluding \emph{non-projective} trees, where dependencies are allowed to cross.
For projective parsing, efficient algorithms exist to solve the three problems, for certain factorizations with special structures. Non-projective parsing with high-order factorizations is known to be NP-hard in computation~\cite{McDonald:EACL06,McDonald:iwpt2007}.
In addition, our models capture multi-root trees, whose root-symbols have one or more children.
A multi-root parser is more robust to sentences that contain disconnected but coherent fragments, since it is
allowed to split up its analysis into multiple pieces.

\subsection{Labeled Parsing}
Our probabilistic model are easily extended to include dependency labels. We denote $\mathrm{L}$ as the set of all
valid dependency labels. We change the feature functions to include label function:
\begin{displaymath}
F_{j}(\boldsymbol{y},\boldsymbol{x})=\sum_{(p,l)\in y}f_{j}(p, l, \boldsymbol{x}).
\end{displaymath}
where $l$ is the vector of dependency labels of edges belonging to part $p$. We define the order of $l$ as the number of labels $l$ contains, and denote it as $o(l)$. It should be noted that the order
of $l$ is not necessarily equal to the order of $p$, since $l$ may contain labels of
parts of edges in $p$. For example, for the second-order sibling model and the part $(s,r,t)$,
$l$ can be defined to contain only the label of edge from word $x_{s}$ to word $x_{t}$.

The weight function of each part is changed to:
\begin{equation}\label{equ:lwt}
w(p,l,\boldsymbol{x})=\exp \bigg\{ \sum_{j}\lambda_{j}f_{j}(p, l, \boldsymbol{x}) \bigg\}.
\end{equation}
Based on Equation~\ref{equ:lwt}, Problem 2 and 3 are rewritten as follows:
\begin{displaymath}
Z(\boldsymbol{x})=\sum\limits_{y\in \mathrm{T}(x)}\bigg[\prod\limits_{(p,l)\in y} w(p, l, \boldsymbol{x})\bigg].
\end{displaymath}
and
\begin{displaymath}
m(p,l)=\sum\limits_{y\in \mathrm{T}(p,l,x)}\textrm{Pr}(\boldsymbol{y}|\boldsymbol{x}) \textrm{, for all }(p,l).
\end{displaymath}
This extension increases the computational complexity of time by factor of $O(|\mathrm{L}|^{o(l)})$, where $|\mathrm{L}|$ is the size of $\mathrm{L}$.

\section{Related Work}
\label{sec:rw}
\subsection{Grammatical Bigram Probability Model}
The probabilistic model described in Section~\ref{subsec:pm} is a generalized formulation of the grammatical
bigram probabilistic model proposed in Eisner~\shortcite{eisn:1996}, which is used by several
works~\cite{Paskin:2001,Koo-EMNLP:2007,Smith-EMNLP:2007}. In fact, the grammatical bigram probabilistic
model is a special case of our probabilistic model, by specifying the parts $p$ as individual edges.
The grammatical bigram model is based on a strong independence assumption: that all the dependency edges of a
tree are independent of one another, given the sentence $\boldsymbol{x}$.

For the first-order model (part $p$ is an individual edge), a variant of the inside-outside
algorithm, which was proposed by Baker~\shortcite{Baker:1979} for probabilistic context-free grammars,
can be applied for the computation of partition function and marginals for projective dependency structures.
This inside-outside algorithm is built on the semiring parsing framework~\cite{goodman:1999}.
For non-projective cases, Problems 2 and 3 can be solved by an adaptation of Kirchhoff's Matrix-Tree
Theorem~\cite{Koo-EMNLP:2007,Smith-EMNLP:2007}.

\begin{figure}[t]
\centering
\epsfig{figure=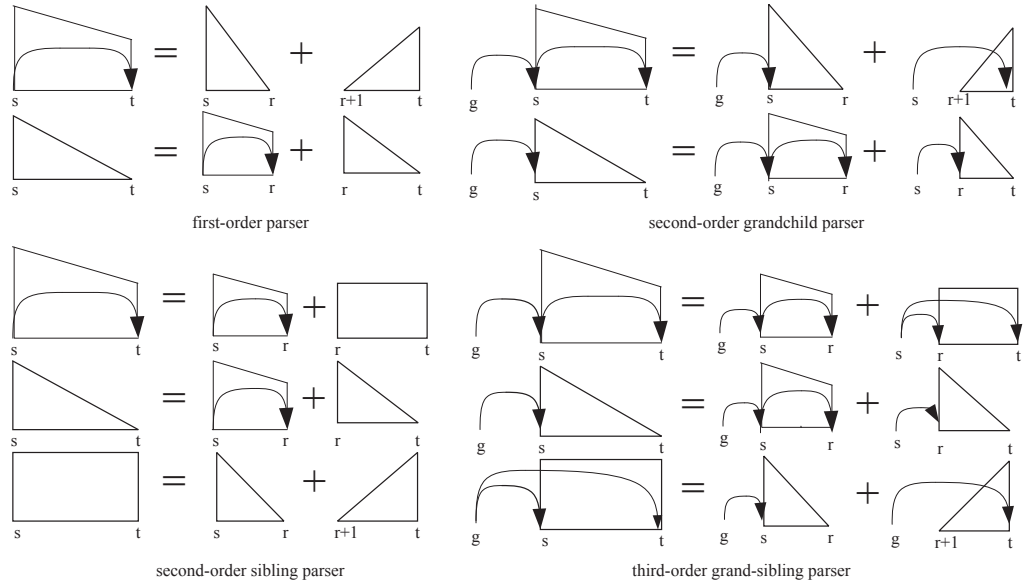,width=\textwidth}
\caption{The dynamic-programming structures and derivation of four graph-based
dependency parsers with different types of factorization.
Symmetric right-headed versions are elided for brevity.}
\label{fig:derv}
\end{figure}

\subsection{Algorithms of Decoding Problem for Different Factored Models}
It should be noted that if the \emph{score} of parts is defined as the logarithm of their weight:
\begin{displaymath}
score(p,\boldsymbol{x})=\log w(p,\boldsymbol{x})=\sum_{j}\lambda_{j}f_{j}(p,\boldsymbol{x}),
\end{displaymath}
then the decoding problem is equivalent to the form of graph-based dependency parsing with global linear model~(GLM), and
several parsing algorithms for different factorizations have been proposed in previous work.
Figure~\ref{fig:derv} provides graphical specifications of these parsing algorithms.

McDonald et al.~\shortcite{McDonald:2005} presented the first-order dependency parser, which decomposes a dependency tree
into a set of individual edges. A widely-used dynamic programming algorithm~\cite{eisner-2000-iwptbook} was used for decoding.
This algorithm introduces two interrelated types of dynamic programming structures: \emph{complete}
spans, and \emph{incomplete} spans \cite{McDonald:2005}. Larger spans are created from
two smaller, adjacent spans by recursive combination in a bottom-up procedure.

The second-order sibling parser~\cite{McDonald:EACL06}
breaks up a dependency tree into \emph{sibling} parts---pairs of adjacent edges with
shared head. Koo and Collins~\shortcite{Koo:2010}
proposed a parser that factors each dependency tree into a set of \emph{grandchild} parts.
Formally, a grandchild part is a triple of indices $(g,s,t)$ where
$g$ is the head of $s$ and $s$ is the head of $t$. In order to parse this
factorization, it is necessary to augment both complete and
incomplete spans with grandparent indices. Following Koo and Collins~\shortcite{Koo:2010},
we refer to these augmented structures as \emph{g-spans}.

The second-order parser proposed in Carreras~\shortcite{cars:2007}
is capable to score both sibling and grandchild parts with complexities of $O(n^{4})$
time and $O(n^{3})$ space. However, the parser suffers an crucial limitation
that it can only evaluate events of grandchild parts for outermost grandchildren.

The third-order grand-sibling parser, which encloses grandchild and sibling parts into a \emph{grand-sibling} part,
was described in Koo and Collins~\shortcite{Koo:2010}. This factorization defines
all grandchild and sibling parts and still requires $O(n^{4})$ time and $O(n^{3})$ space.

\subsection{Transition-based Parsing}
Another category of dependency parsing systems is ``transition-based'' parsing \cite{Nivre:2004,Attardi:conll2006,McDonaldNivre:2007} which parameterizes models over transitions from
one state to another in an abstract state-machine. In these models, dependency trees are constructed by
taking highest scoring transition at each state until a state for the termination is entered.
Parameters in these models are typically learned using standard classification
techniques to predict one transition from a set of possible transitions given a state history.

Recently, several approaches have been proposed to improve transition-based dependency parsers.
In the aspect of decoding, beam search~\cite{Johansson:emnlp07,huang:emnlp09} and partial
dynamic programming~\cite{huang:ACL10} have been applied to improve one-best search.
In the aspect of training, global structural learning has been applied to replace local
learning on each decision~\cite{zhang:2008,huang:emnlp09}.

\section{Algorithms for High-order Models}
\label{sec:al}
In this section, we describe our algorithms for problem 2 and 3 of three high-order factored models: grandchild
and sibling, two second-order models; and grand-sibling, which is third-order. Our algorithms are built on the idea from
the inside-outside algorithm~\cite{Paskin:2001} for the first-order projective parsing model. Following this, we
define the inside probabilities $\beta$ and outside probabilities $\alpha$ over spans $\phi$:
\begin{displaymath}
\begin{array}{lll}
\beta(\phi) & = & \sum\limits_{t\in \phi}\prod\limits_{p\in t} w(p,\boldsymbol{x}) \\
\alpha(\phi) & = & \sum\limits_{y\in \mathrm{T}(\phi)}\prod\limits_{p\not\in y(\phi)} w(p,\boldsymbol{x}),
\end{array}
\end{displaymath}
where $t$ is a sub-structure of a tree and $\boldsymbol{y}(\phi)$ is the sub-structure of
tree $\boldsymbol{y}$ that belongs to span $\phi$.

\begin{table}[t]
{\renewcommand{\tablename}{Algorithm}
\caption{Compute inside probability $\beta$ for second-order Grandchild Model} \label{algm1}
\begin{algorithm}
\item[Require] $\beta(C_{s,s}^{g}) = 1.0 \quad \forall g, s$
\item[1] \textbf{for }$k=1$ to $n$
\item[2] $\textrm{} \quad$\textbf{for }$s=0$ to $n-k$
\item[3] $\textrm{} \qquad t = s + k$
\item[4] $\textrm{} \qquad$\textbf{for }$g < s$ or $g > t$
\item[5] $\textrm{} \quad \textrm{} \quad \beta(I_{s,t}^{g}) = \sum\limits_{s \leq r < t}\beta(C_{s,r}^{g}) \cdot \beta(C_{t,r+1}^{s}) \cdot w_{s,t}^{g} \quad \beta(I_{t,s}^{g}) = \sum\limits_{s \leq r < t}\beta(C_{s,r}^{t}) \cdot \beta(C_{t,r+1}^{g}) \cdot w_{t,s}^{g}$
\item[6] $\textrm{} \quad \textrm{} \quad \beta(C_{s,t}^{g}) = \sum\limits_{s < r \leq t}\beta(I_{s,r}^{g}) \cdot \beta(C_{r,t}^{s}) \qquad \textrm{} \qquad \,\, \beta(C_{t,s}^{g}) = \sum\limits_{s \leq r < t}\beta(I_{t,r}^{g}) \cdot \beta(C_{r,s}^{t})$
\item[7] $\textrm{} \quad$\textbf{end for}
\item[8] \textbf{end for}
\item[Require] $\beta(C_{s,s}) = 1.0 \quad \forall s$
\item[9] $\,\,$ \textbf{for }$k=1$ to $n$
\item[10] $\textrm{} \quad s = n-k, t = k$
\item[11] $\textrm{} \quad \beta(I_{0,t}) = \sum\limits_{0 \leq r < t}\beta(C_{0,r}) \cdot \beta(C_{t,r+1}^{0}) \cdot w_{0,t}^{0} \,\,\,\,\,\, \beta(I_{n,s}) = \sum\limits_{s \leq r < n}\beta(C_{s,r}^{n}) \cdot \beta(C_{n,r+1}) \cdot w_{n,s}^{n}$
\item[12] $\textrm{} \quad \beta(C_{0,t}) = \sum\limits_{0 < r \leq t}\beta(I_{0,r}) \cdot \beta(C_{r,t}^{0}) \qquad \textrm{} \qquad \,\, \beta(C_{n,s}) = \sum\limits_{s \leq r < n}\beta(I_{n,r}) \cdot \beta(C_{r,s}^{n})$
\item[13] \textbf{end for}
\end{algorithm}
}
\end{table}

\begin{table}[!t]
{\renewcommand{\tablename}{Algorithm}
\caption{Compute outside probability $\alpha$ for second-order Grandchild Model} \label{algm2}
{\small
\begin{algorithm}
\item[Require] $\alpha(I_{0,n}) = 1.0$, $\alpha(I_{n,0}) = 1.0$
\item[1] \textbf{for }$k=n$ to $1$
\item[2] $\textrm{} \quad s = n-k, t = k$
\item[3] $\textrm{} \quad \alpha(C_{0,t}) = \sum\limits_{t < r \leq n}\beta(C_{r,t+1}^{0}) \cdot \alpha(I_{0,r}) \cdot w_{0,r}^{0} \qquad \textrm{} \qquad \,\, \alpha(C_{n,s}) = \sum\limits_{0 \leq r < s}\beta(C_{r,s-1}^{n}) \cdot \alpha(I_{n,r}) \cdot w_{n,r}^{n}$
\item[4] $\textrm{} \quad \alpha(I_{0,t}) = \sum\limits_{t \leq r \leq n}\beta(C_{t,r}^{0}) \cdot \alpha(C_{0,r}) \qquad \textrm{} \qquad \textrm{} \qquad \textrm{} \qquad \alpha(I_{n,s}) = \sum\limits_{0 \leq r \leq s}\beta(C_{s,r}^{n}) \cdot \alpha(C_{n,r})$
\item[5] \textbf{end for}
\item[Require] $\alpha(I_{0,n}^{0}) = 1.0$, $\alpha(I_{n,0}^{n}) = 1.0$
\item[6] \textbf{for }$k=n$ to $1$
\item[7] $\textrm{} \quad$\textbf{for }$s=0$ to $n-k$
\item[8] $\textrm{} \qquad t = s + k$
\item[9] $\textrm{} \qquad$\textbf{for }$g < s$
\item[10] $\textrm{} \qquad \textrm{} \quad \alpha(C_{s,t}^{g}) = \sum\limits_{t < r \leq n}\beta(C_{r,t+1}^{s}) \cdot \alpha(I_{s,r}^{g}) \cdot w_{s,r}^{g} + \sum\limits_{r < g \lor r > t}\beta(I_{g,s}^{r}) \cdot \alpha(C_{g,t}^{r})$
\item[11] $\textrm{} \qquad \textrm{} \quad \alpha(C_{t,s}^{g}) = \sum\limits_{g < r < s}\beta(C_{r,s-1}^{t}) \cdot \alpha(I_{t,r}^{g}) \cdot w_{t,r}^{g} + \sum\limits_{r < g \lor r > t}\beta(C_{g,s-1}^{r}) \cdot \alpha(I_{g,t}{r}) \cdot w_{g,t}^{r}$
\item[12] $\textrm{} \qquad \textrm{} \quad$\textbf{if }$g=0$
\item[13] $\textrm{} \qquad \textrm{} \qquad\alpha(C_{s,t}^{g}) \stackrel{+}{=} \beta(I_{0,s}) \cdot \alpha(C_{0,t}) \qquad
    \textrm{} \qquad \textrm{} \qquad \textrm{} \quad \,\,\,\, \alpha(C_{t,s}^{g}) \stackrel{+}{=} \beta(C_{0,s-1}) \cdot \alpha(I_{0,t}) \cdot w_{0,t}^{0}$
\item[14] $\textrm{} \qquad \textrm{} \quad$\textbf{end if}
\item[15] $\textrm{} \qquad \textrm{} \quad \alpha(I_{s,t}^{g}) = \sum\limits_{t \leq r \leq n} \beta(C_{t,r}^{s}) \cdot \alpha(C_{s,r}^{g}) \qquad \textrm{} \qquad \textrm{} \qquad \,\, \alpha(I_{t,s}^{g}) = \sum\limits_{g < r \leq s} \beta(C_{s,r}^{t} \cdot \alpha(C_{t,r}^{g})$
\item[16] $\textrm{} \qquad$\textbf{end for}
\item[17] $\textrm{} \qquad$\textbf{for }$g > t$
\item[18] $\textrm{} \qquad \textrm{} \quad \alpha(C_{s,t}^{g}) = \sum\limits_{t < r < g}\beta(C_{r,t+1}^{s}) \cdot \alpha(I_{s,r}^{g}) \cdot w_{s,r}^{g} + \sum\limits_{r < s \lor r > g}\beta(C_{g,t+1}^{r}) \cdot \alpha(I_{g,s}^{r}) \cdot w_{g,s}^{r}$
\item[19] $\textrm{} \qquad \textrm{} \quad \alpha(C_{t,s}^{g}) = \sum\limits_{0 leq r < s}\beta(C_{r,s-1}^{t}) \cdot \alpha(I_{t,r}^{g}) \cdot w_{t,r}^{g} + \sum\limits_{r < s \lor r > g}\beta(I_{g,t}^{r}) \cdot \alpha(I_{g,s}{r})$
\item[20] $\textrm{} \qquad \textrm{} \quad$\textbf{if }$g=n$
\item[21] $\textrm{} \qquad \textrm{} \qquad \alpha(C_{s,t}^{g}) \stackrel{+}{=} \beta(I_{n,t+1}) \cdot \alpha(C_{n,s}) \cdot w_{n,s}^{n} \qquad \textrm{} \qquad \alpha(C_{t,s}^{g}) \stackrel{+}{=} \beta(I_{n,t}) \cdot \alpha(I_{n,s})$
\item[22] $\textrm{} \qquad \textrm{} \quad$\textbf{end if}
\item[23] $\textrm{} \qquad \textrm{} \quad \alpha(I_{s,t}^{g}) = \sum\limits_{t \leq r < g} \beta(C_{t,r}^{s}) \cdot \alpha(C_{s,r}^{g}) \qquad \textrm{} \qquad \textrm{} \qquad \,\, \alpha(I_{t,s}^{g}) = \sum\limits_{0 \leq r \leq s} \beta(C_{s,r}^{t}) \cdot \alpha(C_{t,r}^{g})$
\item[24] $\textrm{} \qquad$\textbf{end for}
\item[25] $\textrm{} \quad$\textbf{end for}
\item[26] \textbf{end for}
\end{algorithm}
}}
\end{table}

\subsection{Model of Grandchild Factorization}
In the second-order grandchild model, each dependency tree is factored into a set of \emph{grandchild} parts---
pairs of dependencies connected head-to-tail. Formally, a grandchild part is a triple of indices $(g,s,t)$
where both $(g,s)$ and $(s,t)$ are dependencies.

In order to compute the partition function $Z(\boldsymbol{x})$ and marginals $m(g,s,t)$ for this factorization, we
augment both incomplete and complete spans with grandparent indices. This is similar to
Koo and Collins~\shortcite{Koo:2010} for the decoding algorithm of this grandchild factorization.
Following Koo and Collins~\shortcite{Koo:2010}, we refer to these augmented structures as \emph{g-spans}, and
denote an incomplete g-span as $I_{s,t}^{g}$, where $I_{s,t}$ is a normal complete span and $g$ is the index of a
grandparent lying outside the range $[s,t]$, with the implication that $(g,s)$ is a dependency. Complete g-spans are
defined analogously and denoted as $C_{s,t}^{g}$. In addition, we denote the weight of a grandchild part $(g,s,t)$ as
$w_{s,t}^{g}$ for brevity.

The algorithm for the computation of inside probabilities $\beta$ is shown as Algorithm \ref{algm1}. The dynamic
programming derivations resemble those of the decoding algorithm of this factorization,
the only difference is to replace the maximization with summation.
The reason is obvious, since the spans defined for the two algorithms are the same. Note that since our
algorithm considers multi-root dependency trees, we should perform another recursive step to compute the inside
probability $\beta$ for the complete span $C_{0,t}$, after the computation of $\beta$ for all g-spans.

Algorithm \ref{algm2} illustrates the algorithm for computing outside probabilities $\alpha$. This is a top-down
dynamic programming algorithm, and the key of this algorithm is to determine all the contributions to the final
$Z(\boldsymbol{x})$ for each g-span; fortunately, this can be done deterministically for all cases. For example,
the complete g-span $C_{s,t}^{g}$ with $g<s<t$ has two different contributions: combined with a g-span $C_{r,t+1}^{s}$,
of which $r > t$, in the right side to build up a larger g-span $I_{s,r}^{g}$; or combined with a g-span $I_{g,s}^{r}$,
of which $r > t$ or $r < g$, in the left side to form a larger g-span $C_{g,t}^{r}$. So $\alpha(C_{s,t}^{g})$
is the sum of two items, each of which corresponds to one of the two cases (See Algorithm~\ref{algm2}). It should be noted that
complete g-spans $C_{s,t}^{g}$ with $g=0$ or $g=n$ are two special cases.

After the computation of $\beta$ and $\alpha$ for all spans, we can get marginals using following equation:
\begin{displaymath}
m(g,s,t)=\beta(I_{s,t}^{g}) \cdot \alpha(I_{s,t}^{g}) / z(\boldsymbol{x}).
\end{displaymath}
Since the complexity of the both Algorithm \ref{algm1} and Algorithm \ref{algm2} is $O(n^{4})$ time and $O(n^{3})$
space, the complexity overall for training this model is $O(n^{4})$ time and $O(n^{3})$ space, which is the same as
the decoding algorithm of this factorization.


\subsection{Model of Sibling Factorization}
In order to parse the sibling factorization, a new type of span: \emph{sibling} spans, is defined~\cite{McDonald:2006}.
We denote a sibling span as $S_{s,t}$ where $s$ and $t$ are successive modifiers with a shared head.
Formally, a sibling span $S_{s,t}$ represents the region between successive modifiers $s$ and $t$ of some head.
The graphical specification of the second-order sibling model for dynamic-programming, which is in the original work of
Eisner~\cite{eisn:1996}, is shown in Figure~\ref{fig:derv}.
The key insight is that an incomplete span is constructed by combining a smaller incomplete span with a sibling span
that covers the region between the two successive modifiers. The new way allows for the collection of pairs of sibling
dependents in a single state.
It is no surprise that the dynamic-programming structures and derivations of the algorithm for
computing $\beta$ is the same as that of the decoding algorithm, and we omit the pseudo-code of this algorithm.

The algorithm for computing $\alpha$ can be designed with the new dynamic programming structures. The pseudo-code of this
algorithm is illustrated in Algorithm \ref{algm3}. We use $w_{s,r,t}$ to denote the weight of a sibling part $(s,r,t)$.
The computation of marginals of sibling parts is quite different from that of the first-order dependency or second-order
grandchild model. For the introduction of sibling spans, two different cases should be considered:
the modifiers are at the left/right side of the head. In addition, the part $(s,-,t)$, which represents that
$t$ is the inner-most modifier of $s$, is a special case and should be
treated specifically. We can get marginals for all sibling parts with $s < r < t$ as following:
\begin{displaymath}
{\setlength\arraycolsep{2pt}
\begin{array}{lll}
m(s,r,t) & = & \beta(I_{s,r}) \cdot \beta(S_{r,t}) \cdot \alpha(I_{s,t}) \cdot w_{s,r,t} / z(\boldsymbol{x}) \\
m(t,r,s) & = & \beta(S_{s,r}) \cdot \beta(I_{t,r}) \cdot \alpha(I_{t,s}) \cdot w_{t,r,s} / z(\boldsymbol{x}) \\
m(s,-,t) & = & \beta(C_{t,s+1}) \cdot \alpha(I_{s,t}) \cdot w_{s,-,t} / z(\boldsymbol{x}) \\
m(t,-,s) & = & \beta(C_{s,t-1}) \cdot \alpha(I_{t,s}) \cdot w_{t,-,s} / z(\boldsymbol{x}),
\end{array}
}
\end{displaymath}
Since each derivation is defined by a span and a split point, the complexity for training and
decoding of the second-order sibling model is $O(n^{3})$ time and $O(n^{2})$ space.

\begin{table}[t]
{\renewcommand{\tablename}{Algorithm}
\caption{Compute outside probability $\alpha$ for second-order Sibling Model}\label{algm3}
\begin{algorithm}
\item[Require] $\alpha(C_{0,n}) = 1.0 \quad \alpha(C_{n,0}) = 1.0$
\item[1] \textbf{for }$k=n$ to $1$
\item[2] $\textrm{} \quad$\textbf{for }$s=0$ to $n-k$
\item[3] $\textrm{} \qquad t = s + k$
\item[4] $\textrm{} \qquad \alpha(S_{s,t}) = \sum\limits_{0 \leq r < s}\beta(I_{r,s}) \cdot \alpha(I_{r,t}) \cdot w_{r,s,t} + \sum\limits_{t < r \leq n}\beta(I_{r,t}) \cdot \alpha(I_{r,s}) \cdot w_{r,t,s}$
\item[5] $\textrm{} \qquad \alpha(C_{s,t}) = \sum\limits_{t < r \leq n}\beta(C_{r,t+1}) \cdot \alpha(S_{s,r}) + \sum\limits_{0 \leq r < s}\beta(I_{r,s}) \cdot \alpha(C_{r,t})$
\item $\textrm{} \qquad \textrm{} \qquad \textrm{} \qquad + \beta(C_{t+1,t+1}) \cdot \alpha(I_{t+1,s}) \cdot w_{t+1,-,s}$
\item[6] $\textrm{} \qquad \alpha(C_{t,s}) = \sum\limits_{0 \leq r < s}\beta(C_{r,s-1}) \cdot \alpha(S_{r,t})
    + \sum\limits_{t < r \leq n}\beta(I_{r,t}) \cdot \alpha(C_{r,s})$
\item $\textrm{} \qquad \textrm{} \qquad \textrm{} \qquad + \beta(C_{s-1,s-1}) \cdot \alpha(I_{s-1,t}) \cdot w_{s-1,-,t}$
\item[7] $\textrm{} \qquad \alpha(I_{s,t}) = \sum\limits_{t < r \leq n}\beta(S_{t,r}) \cdot \alpha(I_{s,r}) \cdot w_{s,t,r}
    + \sum_{t \leq r \leq n}\beta(C_{r,t}) \cdot \alpha(C_{s,r})$
\item[8] $\textrm{} \qquad \alpha(I_{t,s}) = \sum_{0 \leq r < s}\beta(S_{r,s}) \cdot \alpha(I_{t,r}) \cdot w_{t,s,r} + \sum_{0 \leq r \leq s}\beta(C_{s,r}) \cdot \alpha(C_{t,r})$
\item[9] $\textrm{} \quad$\textbf{end for}
\item[10] \textbf{end for}
\end{algorithm}
}
\end{table}

\begin{table}[!p]
{\renewcommand{\tablename}{Algorithm}
\caption{Compute outside probability $\alpha$ for third-order Grand-sibling Model} \label{algm4}
{\footnotesize
\begin{algorithm}
\item[Require] $\alpha(I_{0,n}) = 1.0$, $\alpha(I_{n,0}) = 1.0$, $\alpha(C_{0,n}) = 1.0$, $\alpha(C_{n,0}) = 1.0$
\item[1] \textbf{for }$k=n$ to $1$
\item[2] $\textrm{} \quad s = n-k, t = k$
\item[3] $\textrm{} \quad \alpha(I_{0,t}) = \beta(C_{t,n}^{0}) \cdot \alpha(C_{0,n}) + \sum\limits_{t < r \leq n}\beta(S_{r,t}^{0}) \cdot \alpha(I_{0,r}) \cdot w_{0,t,r}^{0}$
\item[4] $\textrm{} \quad \alpha(I_{n,s}) = \beta(C_{0,s}^{n}) \cdot \alpha(C_{n,0}) + \sum\limits_{0 \leq r < s}\beta(S_{r,s}^{n}) \cdot \alpha(I_{n,r}) \cdot w_{n,s,r}^{n}$
\item[5] \textbf{end for}
\item[Require] $\alpha(I_{0,n}^{0}) = 1.0$, $\alpha(I_{n,0}^{n}) = 1.0$
\item[6] \textbf{for }$k=n$ to $1$
\item[7] $\textrm{} \quad$\textbf{for }$s=0$ to $n-k$
\item[8] $\textrm{} \qquad t = s + k$
\item[9] $\textrm{} \qquad$\textbf{for }$g < s$
\item[10] $\textrm{} \qquad \textrm{} \quad \alpha(S_{s,t}^{g}) = \sum\limits_{r < g \lor r > t}\beta(I_{g,s}^{r}) \cdot \alpha(I_{g,t}^{r}) \cdot w_{g,s,t}^{r} \qquad \textrm{} \qquad \textbf{if } g = 0 \quad \alpha(S_{s,t}^{g}) \stackrel{+}{=} \beta(I_{0,s}) \cdot \alpha(I_{0,t}) \cdot w_{0,s,t}^{0}$
\item[11] $\textrm{} \qquad \textrm{} \quad \alpha(C_{s,t}^{g}) = \sum\limits_{t < r \leq n}\beta(C_{r,t+1}^{g}) \cdot \alpha(S_{s,r}^{g}) \quad + \sum\limits_{r < g \lor r > t}\beta(I_{g,s}^{r}) \cdot \alpha(C_{g,t}^{r})$
\item[12] $\textrm{} \qquad \textrm{} \quad \alpha(C_{t,s}^{g}) = \sum\limits_{g < r < s}\beta(C_{r,s-1}^{g}) \cdot \alpha(S_{r,t}^{g})
    \quad \textbf{if } g = s - 1 \quad \alpha(C_{t,s}^{g}) \stackrel{+}{=} \sum\limits_{r < g \lor r > t}\beta(C_{s,s}^{g})
    \cdot \alpha(I_{g,t}^{r}) \cdot w_{g,-,t}^{r}$
\item[13] $\textrm{} \qquad \textrm{} \quad \alpha(I_{s,t}^{g}) = \sum\limits_{t < r \leq n}\beta(S_{t,r}^{s}) \cdot \alpha(I_{s,r}^{g}) \cdot w_{s,t,r}^{g} + \sum\limits_{t \leq r \leq n}\beta(C_{t,r}^{s}) \cdot \alpha(C_{s,r}^{g})$
\item[14] $\textrm{} \qquad \textrm{} \quad \alpha(I_{t,s}^{g}) = \sum\limits_{g < r < s}\beta(S_{r,s}^{t}) \cdot \alpha(I_{t,r}^{g}) \cdot w_{t,s,r}^{g} + \sum\limits_{g < r \leq s}\beta(C_{s,r}^{t}) \cdot \alpha(C_{t,r}^{g})$
\item[15] $\textrm{} \qquad$\textbf{end for}
\item[16] $\textrm{} \qquad$\textbf{for }$g > t$
\item[17] $\textrm{} \qquad \textrm{} \quad \alpha(S_{s,t}^{g}) = \sum\limits_{r < s \lor r > g}\beta(I_{g,t}^{r}) \cdot \alpha(I_{g,s}^{r}) \cdot w_{g,t,s}^{r} \qquad \textrm{} \qquad \textbf{if } g = n \quad \alpha(S_{s,t}^{g}) \stackrel{+}{=} \beta(I_{n,t}) \cdot \alpha(I_{n,s}) \cdot w_{n,t,s}^{n}$
\item[18] $\textrm{} \qquad \textrm{} \quad \alpha(C_{s,t}^{g}) = \sum\limits_{t < r < g}\beta(C_{r,t+1}^{g}) \cdot \alpha(S_{s,r}^{g})
    \quad \textbf{if } g = t + 1 \quad \alpha(C_{s,t}^{g}) \stackrel{+}{=} \sum\limits_{r < s \lor r > g}\beta(C_{t,t}^{g})
    \cdot \alpha(I_{s,g}^{r}) \cdot w_{g,-,s}^{r}$
\item[19] $\textrm{} \qquad \textrm{} \quad \alpha(C_{t,s}^{g}) = \sum\limits_{0 \leq r < s}\beta(C_{r,s-1}^{g}) \cdot \alpha(S_{r,t}^{g}) \quad + \sum\limits_{r < s \lor r > g}\beta(I_{g,t}^{r}) \cdot \alpha(C_{g,s}^{r})$
\item[20] $\textrm{} \qquad \textrm{} \quad \alpha(I_{s,t}^{g}) = \sum\limits_{t < r < g}\beta(S_{t,r}^{s}) \cdot \alpha(I_{s,r}^{g}) \cdot w_{s,t,r}^{g} + \sum\limits_{t \leq r < g}\beta(C_{t,r}^{s}) \cdot \alpha(C_{s,r}^{g})$
\item[21] $\textrm{} \qquad \textrm{} \quad \alpha(I_{t,s}^{g}) = \sum\limits_{0 \leq r < s}\beta(S_{r,s}^{t}) \cdot \alpha(I_{t,r}^{g}) \cdot w_{t,s,r}^{g} + \sum\limits_{0 \leq r \leq s}\beta(C_{s,r}^{t}) \cdot \alpha(C_{t,r}^{g})$
\item[22] $\textrm{} \qquad$\textbf{end for}
\item[23] $\textrm{} \quad$\textbf{end for}
\item[24] \textbf{end for}
\end{algorithm}
}}
\end{table}

\subsection{Model of Grand-Sibling Factorization}
We now describe the algorithms of the third-order grand-sibling model.
In this model, each tree is decomposed into \emph{grand-sibling} parts, which enclose grandchild and sibling parts.
Formally, a grand-sibling is a 4-tuple of indices $(g,s,r,t)$ where $(s,r,t)$ is a sibling part and $(g,s,t)$ is a
grandchild part. The algorithm of this factorization can be designed based on the algorithms for grandchild and sibling
models.

Like the extension of the second-order sibling model to the first-order dependency model, we define the sibling g-spans
$S_{s,t}^{g}$, where $S_{s,t}$ is a normal sibling span and $g$ is the index of the head of $s$ and $t$,
which lies outside the region $[s,t]$ with the implication that $(g,s,t)$ forms a valid sibling part.
This model can also be treated as an extension of the sibling model by augmenting it with a
grandparent index for each span, like the behavior of the grandchild model for the first-order dependency model.
Figure~\ref{fig:derv} provides the graphical specification of this factorization for dynamic-programming, too.
The overall structures and derivations is similar to the second-order sibling model, with the addition of grandparent indices.
The same to the second-order grandchild model, the grandparent indices can be set deterministically in all cases.

The pseudo-code of the algorithm for the computation of the outside probability $\alpha$ is illustrated in Algorithm~\ref{algm4}.
It should be noted that in this model there are two types of special cases---one is the sibling-g-span $S_{s,t}^{g}$ with
$g=0$ or $g=n$, as the complete g-span $C_{s,t}^{g}$ with $g=0$ or $g=n$ in the second-order grandchild model; another is
the inner-most modifier case as the second-order sibling model.
We use $w_{s,r,t}^{g}$ to denote the weight of a grand-sibling part $(g,s,r,t)$ and the marginals for all grand-sibling parts
with $s < r < t$ can be computed as follows:
\begin{displaymath}
{\setlength\arraycolsep{2pt}
\begin{array}{lll}
m(g,s,r,t) & = & \beta(I_{s,r}^{g}) \cdot \beta(S_{r,t}^{s}) \cdot \alpha(I_{s,t}^{g}) \cdot w_{s,r,t}^{g} / z(\boldsymbol{x}) \\
m(g,t,r,s) & = & \beta(S_{s,r}^{t}) \cdot \beta(I_{t,r}^{g}) \cdot \alpha(I_{t,s}^{g}) \cdot w_{t,r,s}^{g} / z(\boldsymbol{x}) \\
m(g,s,-,t) & = & \beta(C_{t,s+1}^{s}) \cdot \alpha(I_{s,t}^{g}) \cdot w_{s,-,t}^{g} / z(\boldsymbol{x}) \\
m(g,t,-,s) & = & \beta(C_{s,t-1}^{t}) \cdot \alpha(I_{t,s}^{g}) \cdot w_{t,-,s}^{g} / z(\boldsymbol{x}),
\end{array}
}
\end{displaymath}
Despite the extension to third-order parts, each derivation is still defined by a g-span and a split point as
in second-order grandchild model, so training and decoding of the grand-sibling model
still requires $O(n^{4})$ time and $O(n^{3})$ space.

\setcounter{table}{0}

\begin{table}[t]
\centering
\caption{Training, development and test data for
PTB, CTB and PDT. $\# sentences$ and $\# words$ refer to the number of sentences and the number of
words excluding punctuation in each data set, respectively.}
\label{tab:data}
\begin{tabular}[t]{|c|cccc|}
\hline
 & & sections & $\# sentences$ & $\# words$ \\
\hline
 & Training & 2-21 & 39,832 & 843,029 \\
PTB & Dev & 22 & 1,700 & 35,508 \\
 & Test & 23 & 2,416 & 49,892 \\
\hline
 & Training & 001-815; 1001-1136 & 16,079 & 370,777 \\
CTB & Dev & 886-931; 1148-1151 & 804 & 17,426 \\
 & Test & 816-885; 1137-1147 & 1,915 & 42.773 \\
\hline
 & Training & - & 73088 & 1,255,590 \\
PDT & Dev & - & 7,318 & 126,028 \\
 & Test & - & 7,507 & 125,713 \\
\hline
\end{tabular}
\end{table}

\section{Experiments for Dependency Parsing}
\label{sec:pe}
\subsection{Data Sets}
We implement and evaluate the proposed algorithms of the three factored models~(sibling, grandchild and grand-sibling)
on the Penn English Treebank~(PTB version 3.0)~\cite{Marcus:1993},
the Penn Chinese Treebank~(CTB version 5.0)~\cite{Xue:2005} and Prague Dependency Treebank~(PDT)~\cite{Haj:1998,Haj:2001}.

For English, the PTB data is prepared by using the standard split:
sections 2-21 are used for training, section 22 is for development,
and section 23 for test. Dependencies are extracted by using
Penn2Malt\footnote{http://w3.msi.vxu.se/\~{}nivre/research/Penn2Malt.html}
tool with standard head rules~\cite{Yamada:2003}. For Chinese, we adopt the data split
from Zhang and Clark \shortcite{zhang:2009}, and we also used the Penn2Malt tool to convert
the data into dependency structures. Since the dependency trees for English and Chinese are extracted
from phrase structures in Penn Treebanks, they contain no crossing edges by construction.
For Czech, the PDT has a predefined training, developing and testing split. we "projectivized" the training data by finding best-match projective trees\footnote{Projective trees for training
sentences are obtained by running the first-order projective parser with an oracle model that assigns a score of +1 to
correct edges and -1 otherwise.}.

All experiments were running using every single sentence in each set of data regardless of length.
Parsing accuracy is measured with unlabeled attachment score~(UAS): the percentage of
words with the correct head, root accuracy~(RA): the percentage of correctly identified root words,
and the percentage of complete matches~(CM).
Following the standard of previous work, we did not include
punctuation\footnote{English evaluation ignores any token whose gold-standard POS is one of $\{ \textrm{'' `` : , .}\}$;
Chinese evaluation ignores any token whose tag is ``PU''}
in the calculation of accuracies for English and Chinese.
The detailed information of each treebank is showed in Table~\ref{tab:data}.

\begin{table}[t]
\centering
\caption{All feature templates of different factorizations used by our parsing algorithms.
L($\cdot$) and P($\cdot$) are the lexicon and POS tag of each token.}
\label{tab:feat}
\begin{tabular}[t]{|l|l|l|l|}
\hline
\multicolumn{4}{|c|}{\textbf{dependency features for part} $(s,t)$} \\
\hline
\multicolumn{1}{|c|}{\textbf{uni-gram features}} & \multicolumn{2}{c}{\textbf{bi-gram features}} &
\multicolumn{1}{|c|}{\textbf{context features}} \\
\hline
L(s)$\cdot$P(s) & \multicolumn{2}{l|}{L(s)$\cdot$P(s)$\cdot$L(t)$\cdot$P(t)} &
P(s)$\cdot$P(t)$\cdot$P(s+1)$\cdot$P(t-1) \\
L(s) & \multicolumn{1}{l}{L(s)$\cdot$P(s)$\cdot$P(t)} &
\multicolumn{1}{l|}{$\textrm{}\qquad\textrm{}\quad$P(s)$\cdot$L(t)$\cdot$P(t)} &
P(s)$\cdot$P(t)$\cdot$P(s-1)$\cdot$P(t-1) \\
P(s) & \multicolumn{1}{l}{L(s)$\cdot$P(s)$\cdot$L(t)} &
\multicolumn{1}{l|}{$\textrm{}\qquad\textrm{}\quad$L(s)$\cdot$L(t)$\cdot$P(t)} &
P(s)$\cdot$P(t)$\cdot$P(s+1)$\cdot$P(t+1) \\
L(t)$\cdot$P(t) & \multicolumn{1}{l}{L(s)$\cdot$L(t)} &
\multicolumn{1}{l|}{$\textrm{}\qquad\textrm{}\quad$P(s)$\cdot$P(t)} &
P(s)$\cdot$P(t)$\cdot$P(s+1)$\cdot$P(t-1) \\
\cline{2-3}
L(t) & \multicolumn{2}{c|}{\textbf{in between features}} & \\
\cline{2-3}
P(t) & \multicolumn{1}{l}{L(s)$\cdot$L(b)$\cdot$L(t)} &
\multicolumn{1}{l|}{$\textrm{}\qquad\textrm{}\quad$P(s)$\cdot$P(b)$\cdot$P(t)} & \\
\hline
\multicolumn{2}{|c|}{\textbf{grandchild features for part} $(g,s,t)$} &
\multicolumn{2}{c|}{\textbf{sibling features for part} $(s,r,t)$} \\
\hline
\textbf{tri-gram features} & \textbf{backed-off features} & \textbf{tri-gram features} & \textbf{backed-off features} \\
\hline
L(g)$\cdot$L(s)$\cdot$L(t) & L(g)$\cdot$L(t) & L(s)$\cdot$L(r)$\cdot$L(t) & L(r)$\cdot$L(t) \\
P(g)$\cdot$P(s)$\cdot$P(t) & P(g)$\cdot$P(t) & P(s)$\cdot$P(r)$\cdot$P(t) & P(r)$\cdot$P(t) \\
L(g)$\cdot$P(g)$\cdot$P(s)$\cdot$P(t) & L(g)$\cdot$P(t) & L(s)$\cdot$P(s)$\cdot$P(r)$\cdot$P(t) & L(r)$\cdot$P(t) \\
P(g)$\cdot$L(s)$\cdot$P(s)$\cdot$P(t) & P(g)$\cdot$L(t) & P(s)$\cdot$L(r)$\cdot$P(r)$\cdot$P(t) & P(r)$\cdot$L(t) \\
P(g)$\cdot$P(s)$\cdot$L(t)$\cdot$P(t) & & P(s)$\cdot$P(r)$\cdot$L(t)$\cdot$P(t) & \\
\hline
\multicolumn{4}{|c|}{\textbf{grand-sibling features for part} $(g,s,r,t)$} \\
\hline
\multicolumn{1}{|c|}{\textbf{4-gram features}} & \multicolumn{2}{c}{\textbf{context features}} &
\multicolumn{1}{|c|}{\textbf{backed-off features}} \\
\hline
L(g)$\cdot$P(s)$\cdot$P(r)$\cdot$P(t) &
\multicolumn{2}{l|}{P(g)$\cdot$P(s)$\cdot$P(r)$\cdot$P(t)$\cdot$P(g+1)$\cdot$P(s+1)$\cdot$P(t+1)} &
L(g)$\cdot$P(r)$\cdot$P(t) \\
P(g)$\cdot$L(s)$\cdot$P(r)$\cdot$P(t) &
\multicolumn{2}{l|}{P(g)$\cdot$P(s)$\cdot$P(r)$\cdot$P(t)$\cdot$P(g-1)$\cdot$P(s-1)$\cdot$P(t-1)} &
P(g)$\cdot$L(r)$\cdot$P(t) \\
P(g)$\cdot$P(s)$\cdot$L(r)$\cdot$P(t) &
\multicolumn{2}{l|}{P(g)$\cdot$P(s)$\cdot$P(r)$\cdot$P(t)$\cdot$P(g+1)$\cdot$P(s+1)} &
P(g)$\cdot$P(r)$\cdot$L(t) \\
P(g)$\cdot$P(s)$\cdot$P(r)$\cdot$L(t) &
\multicolumn{2}{l|}{P(g)$\cdot$P(s)$\cdot$P(r)$\cdot$P(t)$\cdot$P(g-1)$\cdot$P(s-1)} &
L(g)$\cdot$L(r)$\cdot$P(t) \\
L(g)$\cdot$L(s)$\cdot$P(r)$\cdot$P(t) &
\multicolumn{2}{l|}{P(g)$\cdot$P(r)$\cdot$P(t)$\cdot$P(g+1)$\cdot$P(r+1)$\cdot$P(t+1)} &
L(g)$\cdot$P(r)$\cdot$L(t) \\
L(g)$\cdot$P(s)$\cdot$L(r)$\cdot$P(t) &
\multicolumn{2}{l|}{P(g)$\cdot$P(r)$\cdot$P(t)$\cdot$P(g+1)$\cdot$P(r-1)$\cdot$P(t-1)} &
P(g)$\cdot$L(r)$\cdot$L(t) \\
L(g)$\cdot$P(s)$\cdot$P(r)$\cdot$L(t) &
\multicolumn{2}{l|}{P(g)$\cdot$P(r)$\cdot$P(g+1)$\cdot$P(r+1)} &
P(g)$\cdot$P(r)$\cdot$P(t) \\
P(g)$\cdot$L(s)$\cdot$L(r)$\cdot$P(t) &
\multicolumn{2}{l|}{P(g)$\cdot$P(r)$\cdot$P(g-1)$\cdot$P(r-1)} & \\
L(g)$\cdot$L(s)$\cdot$P(r)$\cdot$L(t) &
\multicolumn{2}{l|}{P(g)$\cdot$P(t)$\cdot$P(g+1)$\cdot$P(t+1)} & \\
P(g)$\cdot$P(s)$\cdot$L(r)$\cdot$L(t) &
\multicolumn{2}{l|}{P(g)$\cdot$P(t)$\cdot$P(g-1)$\cdot$P(t-1)} & \\
P(g)$\cdot$P(s)$\cdot$P(r)$\cdot$P(t) &
\multicolumn{2}{l|}{P(r)$\cdot$P(t)$\cdot$P(r+1)$\cdot$P(t+1)} & \\
 & \multicolumn{2}{l|}{P(r)$\cdot$P(t)$\cdot$P(r-1)$\cdot$P(t-1)} & \\
\hline
\multicolumn{4}{|c|}{\textbf{coordination features}} \\
\hline
\multicolumn{1}{|l}{L(g)$\cdot$P(s)$\textrm{}\quad$P(g)$\cdot$P(s)} &
\multicolumn{2}{l}{L(g)$\cdot$L(s)$\cdot$L(t)$\textrm{}\qquad\textrm{}\qquad$L(g)$\cdot$P(s)$\cdot$P(t)} &
\multicolumn{1}{l|}{P(g)$\cdot$L(s)$\textrm{}\quad$P(g)$\cdot$L(t)} \\
\multicolumn{1}{|l}{L(g)$\cdot$P(t)$\textrm{}\quad$P(g)$\cdot$P(t)} &
\multicolumn{2}{l}{P(g)$\cdot$L(s)$\cdot$P(t)$\textrm{}\qquad\textrm{}\qquad$P(g)$\cdot$P(s)$\cdot$L(t)} &
\multicolumn{1}{l|}{L(s)$\cdot$P(t)$\textrm{}\quad$P(s)$\cdot$L(t)} \\
\multicolumn{1}{|l}{P(s)$\cdot$P(t)} &
\multicolumn{2}{l}{L(g)$\cdot$L(s)$\cdot$P(t)$\textrm{}\qquad\textrm{}\qquad$L(g)$\cdot$P(s)$\cdot$L(t)} &
\multicolumn{1}{l|}{} \\
\multicolumn{1}{|l}{} &
\multicolumn{2}{l}{P(g)$\cdot$L(s)$\cdot$L(t)$\textrm{}\qquad\textrm{}\qquad$P(g)$\cdot$P(s)$\cdot$P(t)} &
\multicolumn{1}{l|}{} \\
\hline
\end{tabular}
\end{table}

\subsection{Feature Space}
Following previous work for high-order dependency parsing~\cite{McDonald:EACL06,cars:2007,Koo:2010}, higher-order
factored models captures not only features associated with corresponding higher order parts,
but also the features of relevant lower order parts that are enclosed in its factorization.
For example, third-order grand-sibling model evaluates parts for dependencies, siblings, grandchildren and grand-siblings,
so that the feature function of a dependency parse is given by:
\begin{eqnarray}
F(\boldsymbol{y},\boldsymbol{x}) & = & \sum\limits_{(s,t)\in y}f_{dep}(s,t,\boldsymbol{x}) \nonumber \\
 & + & \sum\limits_{(s,r,t)\in y}f_{sib}(s,r,t,\boldsymbol{x}) \nonumber \\
 & + & \sum\limits_{(g,s,t)\in y}f_{gch}(g,s,t,\boldsymbol{x}) \nonumber \\
 & + & \sum\limits_{(g,s,r,t)\in y}f_{gsib}(g,s,r,t,\boldsymbol{x}) \nonumber
\end{eqnarray}
where $f_{dep}$, $f_{sib}$, $f_{gch}$, and $f_{gsib}$ are the feature functions of dependency, sibling,
grandchild, and grand-sibling parts.

First-order dependency features $f_{dep}$, second-order sibling features $f_{sib}$ and second-order grandchild
features $f_{gch}$ are based on feature sets from previous work~\cite{McDonald:2005,McDonald:EACL06,cars:2007},
to which we added lexicalized versions of several features. For instance, our first-order feature set contains
lexicalized ``in-between'' features that recognize word types that occur between the head and modifier words in an
attachment decision, while previous work has defined in-between features only for POS tags.
As another example, the second-order features $f_{sib}$ and $f_{gch}$ contains lexical trigram
features, which also excluded in the feature sets of previous work. The third-order grand-sibling features are
based on Koo and Collins~\cite{Koo:2010}. All feature templates for used in our parsers are outlined
in Table~\ref{tab:feat}.

According to Table~\ref{tab:feat}, several features in our parser depend on part-of-speech~(POS) tags of
input sentences. For English, POS tags are automatically assigned by the SVMTool tagger~\cite{Gim:2004};
For Chinese, we used gold-standard POS tags in CTB. Following Koo and Collins~\shortcite{Koo:2010},
two versions of POS tags are used for any features involve POS: one using is normal POS tags and another is
a coarsened version of the POS tags.\footnote{For English, we used the first two characters, except \url{PRP}
and \url{PRP$}; for Czech, we used the first character of the tag;
for Chinese, we dropped the last character, except \url{PU} and \url{CD}.}

\subsection{Model Training}
Since the log-likelihood $L(\lambda)$ is a convex function, gradient descent methods can be used to search for the global
minimum. The method of parameter estimation for our models is the limited memory BFGS algorithm~(L-BFGS)
\cite{Nash:1991}, with L$2$ regularization. L-BFGS algorithm is widely used for large-scale optimization,
as it combines fast training time with low memory requirement which is especially important for large-scale
optimization problems. Meanwhile, L-BFGS can achieve highly competitive performance.
Development sets are used for tuning the hyper-parameter $C$ which dictates the level of the regularization in the model.

For the purpose of comparison, we also run experiments on graph-based dependency parsers
of the three different factorizations, employing two online learning methods: The $k$-best version of the Margin
Infused Relaxed Algorithm~(MIRA)~\cite{Cram1:2003,Cram2:2003,McDonald:2006} with $k=10$, and
averaged structured perceptron~(AP)~\cite{FreundSchapire:1999,Collins:2002}. Both the two learning methods are used in
previous work for training graph-based dependency parsers and achieved highly competitive parsing
accuracies---$k$-best MIRA is used in McDonald et al.~\shortcite{McDonald:2005},
McDonald and Pereira~\shortcite{McDonald:EACL06}, and McDonald and Nivre~\shortcite{McDonaldNivre:2007},
and AP is used in Carreras~\shortcite{cars:2007} and Koo and Collins~\shortcite{Koo:2010}. Each parser is trained for
10 iterations and selects parameters from the iteration that achieves the highest parsing performance on the development set.

The feature sets were fixed for all three languages. For practical reason, we exclude the sentences containing
more than 100 words in all the training data sets of Czech, English and Chinese in all experiments.

\begin{table}[t]
\centering
\caption{UAS, RA and CM of three factored models: Sib for sibling, Gch for grandchild and GSib for grand-sibling.}
\label{tab:main}
\begin{tabular}{|l|ccc|ccc|ccc|}
\hline
 & \multicolumn{9}{c|}{\textbf{Eng}} \\
\cline{2-10}
 & \multicolumn{3}{c|}{\textbf{L-BFGS}} & \multicolumn{3}{c|}{\textbf{MIRA}} & \multicolumn{3}{c|}{\textbf{AP}} \\
\cline{2-10}
 & UAS & RA & CM & UAS & RA & CM & UAS & RA & CM \\
\hline
Sib & 92.4 & \textbf{95.4} & \textbf{46.4} & \textbf{92.5} & 95.1 & 45.7 & 91.9 & 94.8 & 44.1 \\
Gch & 92.2 & \textbf{94.9} & \textbf{44.6} & \textbf{92.3} & 94.7 & 44.0 & 91.6 & 94.5 & 41.6 \\
GSib & 93.0 & \textbf{96.1} & \textbf{48.8} & 93.0 & 95.8 & 48.3 & 92.4 & 95.5 & 46.6 \\
\hline
 & \multicolumn{9}{c|}{\textbf{Chn}} \\
\cline{2-10}
 & \multicolumn{3}{c|}{\textbf{L-BFGS}} & \multicolumn{3}{c|}{\textbf{MIRA}} & \multicolumn{3}{c|}{\textbf{AP}} \\
\cline{2-10}
 & UAS & RA & CM & UAS & RA & CM & UAS & RA & CM \\
Sib & \textbf{86.3} & \textbf{78.5} & \textbf{35.0} & 86.1 & 77.8 & 34.1 & 84.0 & 74.2 & 31.1 \\
Gch & \textbf{85.5} & \textbf{78.0} & \textbf{33.3} & 85.4 & 77.6 & 31.7 & 83.9 & 74.9 & 29.6 \\
GSib & \textbf{87.2} & \textbf{80.0} & \textbf{37.0} & 87.0 & 79.5 & 35.8 & 85.1 & 77.1 & 32.0 \\
\hline
 & \multicolumn{9}{c|}{\textbf{Cze}} \\
\cline{2-10}
 & \multicolumn{3}{c|}{\textbf{L-BFGS}} & \multicolumn{3}{c|}{\textbf{MIRA}} & \multicolumn{3}{c|}{\textbf{AP}} \\
\cline{2-10}
 & UAS & RA & CM & UAS & RA & CM & UAS & RA & CM \\
Sib & \textbf{85.6} & \textbf{90.8} & \textbf{36.3} & 85.5 & 90.5 & 35.1 & 84.6 & 89.5 & 34.0 \\
Gch & \textbf{86.0} & \textbf{91.8} & \textbf{36.5} & 85.8 & 91.4 & 35.6 & 85.0 & 90.2 & 34.6 \\
GSib & \textbf{87.5} & \textbf{93.2} & \textbf{39.3} & 87.3 & 92.9 & 38.4 & 86.4 & 92.1 & 36.9 \\
\hline
\end{tabular}
\end{table}

\subsection{Results and Analysis}
Table~\ref{tab:main} shows the results of three different factored parsing models trained by three different learning
algorithms on the three treebanks of PTB, CTB and PDT. Our parsing models trained by L-BFGS method
achieve significant improvement on parsing performance of the parsing models trained by AP for all the three treebanks,
and obtain parsing performance competitive with the parsing models trained by MIRA.
For example, for the third-order grand-sibling model, the parsers trained by L-BFGS method improve
the UAS of $0.6\%$ for PTB, $2.1\%$ for CTB and $1.1\%$ for PDT, compared with the parsers trained by AP.
For the parsers trained by MIRA, our parsers achieve the same UAS for PTB, and higher parsing
accuracies~(about $0.2\%$ better) for both CTB and PDT. Moreover, it should be noticed that our algorithms
achieve significant improvement of RA and CM on all three treebanks for the parsers trained by MIRA,
although the parsers trained by L-BFGS and MIRA exhibit no statistically significant different in the
parsing performance of UAS. 

\begin{table}[t]
\centering
\caption{Training time for three models. $\# Core$ refers to the number of cores.}
\label{tab:spd}
\begin{tabular}{|l|c|ccc|}
\hline
 & \textbf{MIRA} & \multicolumn{3}{c|}{\textbf{L-BFGS}} \\
$\# Core$ & 1 & 4 & 10 & 18 \\
\hline
Sib & 33.3h & 27.4h & 10.9h & 6.7h \\
Gch & 160.6h & 146.5h & 59.8h & 22.4h \\
GSib & 300.0h & 277.6h & 115.7h & 72.3h \\
\hline
\end{tabular}
\end{table}

As mentioned above, parallel computation techniques could be applied to our models to speed up parser training.
Table~\ref{tab:spd} lists the average training time for our three models with different number of cores.
According to this table, the training time of our parsers trained by off-line L-BFGS method with more than 10 cores is
much less than the cost of the parsers trained by online learning methods MIRA. We omit the training time of online
learning method AP, since the training times for MIRA and AP are nearly the same according to our experiences.
The reason is that the time for updating parameters, which is the only difference between MIRA and AP,
makes up a very small proportion (less than 10\% ) of the total training time.

\begin{table}[b]
\centering
\caption{Accuracy comparisons of different dependency parsers on PTB, CTB and PDT.}
\label{tab:cmp}
\begin{tabular}{|l|cc|cc|cc|}
\hline
 & \multicolumn{2}{c|}{\textbf{Eng}} & \multicolumn{2}{c|}{\textbf{Chn}} & \multicolumn{2}{c|}{\textbf{Cze}} \\
\cline{2-7}
 & UAS & CM & UAS & CM & UAS & CM \\
\hline
McDonald et al. \shortcite{McDonald:2005} & 90.9 & 36.7 & 79.7 & 27.2 & 84.4 & 32.2 \\
McDonald and Pereira \shortcite{McDonald:EACL06} & 91.5 & 42.1 & 82.5 & 32.6 & 85.2 & 35.9 \\
Zhang and Clark \shortcite{zhang:2008} & 92.1 & 45.4 & 85.7 & 34.4 & - & - \\
Zhang and Nivre \shortcite{zhang:ACL11} & 92.9 & 48.0 & 86.0 & 36.9 & - & - \\
Koo and Collins \shortcite{Koo:2010}, model2 & 92.9 & - & - & - & 87.4 & - \\
Koo and Collins \shortcite{Koo:2010}, model1 & 93.0 & - & - & - & 87.4 & - \\
\textbf{this paper} & \textbf{93.0} & \textbf{48.8} & \textbf{87.2} & \textbf{37.0} & \textbf{87.5} & \textbf{39.3} \\
\hline
Koo et al. \shortcite{Koo-dp:2008}$^{*}$ & 93.2 & - & - & - & 87.1 & - \\
Suzuki et al. \shortcite{Suzuki:2009}$^{*}$ & 93.8 & - & - & - & 88.1 & - \\
Zhang and Clark \shortcite{zhang:2009}$^{*}$ & - & - & 86.6 & 36.1 & - & - \\
\hline
\end{tabular}
\end{table}

\subsection{Comparison with Previous Works}
Table \ref{tab:cmp} illustrates the UAS and CM of related work on PTB, CTB and PDT for comparison.
Our experimental results show an improvement in performance of English and Chinese over the results in
Zhang and Clark~\shortcite{zhang:2008}, which combining graph-based and transition-based dependency parsing
into a single parser using the framework of beam-search, and Zhang and Nivre~\shortcite{zhang:ACL11},
which are based on a transition-based dependency parser with rich non-local features.
For English and Czech, our results are better than the results of the two third-order graph-based
dependency parsers in Koo and Collins~\shortcite{Koo:2010}. The models marked * cannot be compared with our work
directly, as they exploit large amount of additional information that is not used in our models,
whiling our parses obtain results competitive with these works.
For example, Koo et al.~\shortcite{Koo-dp:2008} and Suzuki et al.~\shortcite{Suzuki:2009} make use of unlabeled data,
and the parsing model of Zhang and Clark \shortcite{zhang:2009} utilizes phrase structure annotations.

\section{Conclusion}
\label{sec:con}
In this article, we have described probabilistic models for high-order projective dependency parsing,
obtained by relaxing the independent assumption of the previous grammatical bigram model,
and have presented algorithms for computing partition functions and marginals for three factored
parsing models---second-order sibling and grandchild, and third-order grand-sibling.
Our methods achieve competitive or state-of-the-art performance on three treebanks for languages
of English, Chinese and Czech. By analyzing errors on structural properties of
length factors, we have shown that the parsers trained by online and off-line learning methods have distinctive
error distributions despite having very similar parsing performance of UAS overall.
We have also demonstrated that by exploiting parallel computation techniques, our parsing models can be trained
much faster than those parsers using online training methods.

\bibliographystyle{fullname}
\bibliography{max}
\end{document}